\documentclass[conference]{IEEEtran}
\IEEEoverridecommandlockouts
\usepackage{cite}
\usepackage{amsmath,amssymb,amsfonts}
\usepackage{algorithmic}
\usepackage{graphicx}
\usepackage{textcomp}
\usepackage{xcolor}
\usepackage{url}
\usepackage{algorithm}
\usepackage{algorithmic}
\usepackage[colorlinks,urlcolor=blue,citecolor=blue,linkcolor=blue]{hyperref}\usepackage{caption}
\usepackage{setspace}
\usepackage{subcaption} 
\usepackage{graphicx,multicol}
\usepackage{pgfplots}
\usepackage{multirow}
\usepackage{float} 

\makeatletter
\newcommand{\linebreakand}{%
  \end{@IEEEauthorhalign}
  \hfill\mbox{}\par
  \mbox{}\hfill\begin{@IEEEauthorhalign}
}
\makeatother

\begin{document}
\title{Low-Precision Floating-Point for Efficient On-Board Deep Neural Network Processing\\
\thanks{This work is partly supported by the French Space Agency (CNES), Institut National de Recherche en Informatique et en Automatique (INRIA), and the COMINLabs (10-LABX-0007) {LeanAI} project.}}


\author{\IEEEauthorblockN{Cédric Gernigon}
	\IEEEauthorblockA{Univ. Rennes, Inria, CNRS, IRISA\\ F-35000 Rennes, France\\
Email: cedric.gernigon@inria.fr}
\and
	\IEEEauthorblockN{Silviu-Ioan Filip}
	\IEEEauthorblockA{Univ. Rennes, Inria, CNRS, IRISA\\ F-35000 Rennes, France\\
		Email: silviu.filip@inria.fr}
\and
\IEEEauthorblockN{Olivier Sentieys}
	\IEEEauthorblockA{Univ. Rennes, Inria, CNRS, IRISA\\ F-35000 Rennes, France\\
Email: olivier.sentieys@inria.fr}
\linebreakand
\IEEEauthorblockN{Clément Coggiola}
\IEEEauthorblockA{
Spacecraft techniques, on-board data handling\\
CNES, Toulouse, France \\
Email: clement.coggiola@cnes.fr}
\and
\IEEEauthorblockN{Mickaël Bruno}
\IEEEauthorblockA{
Spacecraft techniques, on-board data handling\\
CNES, Toulouse, France \\
Email: mickael.bruno@cnes.fr}
}

\maketitle

\begin{abstract}
One of the major bottlenecks in high-resolution Earth Observation (EO) space systems is the downlink between the satellite and the ground. Due to hardware limitations, on-board power limitations or ground-station operation costs, there is a strong need to reduce the amount of data transmitted. Various processing methods can be used to compress the data. One of them is the use of on-board deep learning to extract relevant information in the data. However, most ground-based deep neural network parameters and computations are performed using single-precision floating-point arithmetic, which is not adapted to the context of on-board processing.
We propose to rely on quantized neural networks and study how to combine low precision (\emph{mini}) floating-point arithmetic with a Quantization-Aware Training methodology. We evaluate our approach with a semantic segmentation task for ship detection using satellite images from the Airbus Ship dataset. Our results show that 6-bit floating-point quantization for both weights and activations can compete with single-precision without significant accuracy degradation. Using a Thin U-Net 32 model, only a 0.3\% accuracy degradation is observed with 6-bit minifloat quantization (a 6-bit equivalent integer-based approach leads to a 0.5\% degradation). An initial hardware study also confirms the potential impact of such low-precision floating-point designs, but further investigation at the scale of a full inference accelerator is needed before concluding whether they are relevant in a practical on-board scenario.
\end{abstract}


\section{Introduction}
Earth Observation (EO) provides an effective way of exploring the physical, chemical, and biological information related to the Earth. This information collected by EO satellites is widely used in various research fields, especially in relation to the environment, where the measurements made by EO satellites are indispensable. Moreover, these new space applications related to Earth observation produce a huge volume of data extracted from various image and radar sensors.
Transmitting all this data is possible through communication between satellites and ground stations.
However, EO systems are limited by these downlink communications, due to hardware limitations, on-board power constraints or ground-station operation cost, for example. Thus, there is a need to reduce the amount of data to be transmitted through the downlink. While data compression is widely used for size reduction, the idea of transmitting only relevant data through on-board processing has only recently started gaining interest.

Artificial Intelligence (AI), and in particular Deep Learning (DL), is starting to be successfully applied in space applications. However, the inference computation of many models is still mainly performed on ground platforms due to their memory footprint and computational intensity~\cite{Furano2020Towards}.
To mitigate this computational burden and the space-to-ground communication bottleneck, recent research has focused on neural network compression~\cite{choudhary2020comprehensive}. Moreover, various techniques such as pruning~\cite{han2015deep}, weight sharing~\cite{dupuis2020automatic}, distillation~\cite{de2020towards}  and quantization~\cite{hubara2017quantized}, can be used to reduce the computational intensity to make it compatible with on-board processing.

Quantization deals with \emph{how} (what number format(s) and bit widths to use) model parameters (such as weights and bias terms) and activation signals (inputs and outputs to layers in a model) are computed and stored at inference time. Furthermore, reducing precision and adapting the number representation make quantization a particularly effective and practical technique.

In this paper, we investigate how to efficiently use quantization to accelerate Deep Neural Network (DNN) inference for space applications, with a strong focus on semantic segmentation tasks. Towards this goal, we present our ongoing work on the efficient use of low-precision floating-point quantization (so-called~\emph{minifloat}s). Our method consists in adapting a quantized DNN training approach that has so far been mostly used for integer/fixed-point-based quantization.
    
\begin{figure*}[h]
    \centering
    \includegraphics[width=0.9\textwidth]{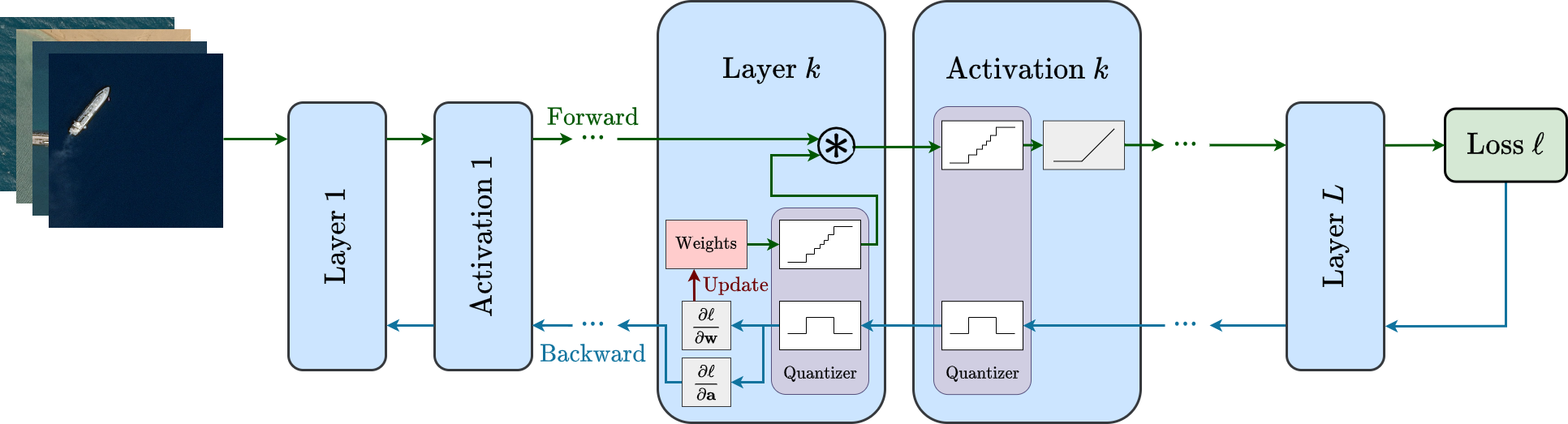}
    \caption{An overview of the QAT scheme we use in the training of minifloat quantized DNNs. The distinguishing feature of a QAT flow compared to a standard training procedure is the presence of quantizer blocks in each layer and activation function that use the STE mechanism to differentiate with respect to the quantized signals.}
    \label{fig:quant}
\end{figure*}

\section{Background and Related Work}

Deep neural network parameters and computations have been traditionally stored and computed using 32-bit single-precision floating-point (FP) arithmetic, making it impossible to deploy state-of-the-art models on low-power and resource-constrained devices without further tuning.

\subsection{DNN Quantization}
Much work in recent years has focused on reducing the bit-width of data and arithmetic in DNN models without impacting task accuracy. Quantization to 8-bit~\cite{jacob_quantization_2018} and sub 8-bit~\cite{zhou_dorefa-net_2018} integer formats has been shown to match single-precision baselines on convolutional networks for several computer vision tasks. In certain cases, it is even possible to go down to extremely low precisions, such as binary~\cite{courbariaux_binarized_nodate} and ternary~\cite{li_ternary_2016} DNNs. 

On the FP arithmetic side, 16-bit formats such as half precision FP16 (5-bit exponent) and bfloat16 (8-bit exponent) have found success in both inference acceleration and mixed-precision DNN training~\cite{kalamkar2019study}. More recently, 8-bit floating-point formats have also been explored in inference/training scenarios~\cite{micikevicius2022fp8, kuzmin2022fp8, wu2020phoenix} and are starting to get hardware support (\emph{e.g.}~the NVIDIA Hopper architecture is an example). Sub 8-bit FP custom formats are also being studied in the literature in the context of model compression and inference acceleration~\cite{tambe2020algorithm}.

To apply such quantization schemes on a DNN model, it is necessary to adjust the parameters according to the target format. There are two main approaches to do this in practice. The first is Post-Training Quantization (PTQ), in which quantization is applied after training. While fast, it can lead to non-negligible loss in accuracy for extremely low bit width formats~\cite{banner_post_nodate}. The other approach is to iteratively quantize the model during training. This process is known as Quantization-Aware Training (QAT). While much slower, it generally leads to better quantization results. For an in-depth overview of quantization in the context of DNNs the reader can consult~\cite{gholami2021survey}.
 
\subsection{Quantization-Aware Training}
The emergence of QAT methods can be traced back to the pioneering work of~\cite{courbariaux_binaryconnect_2015,courbariaux_binarized_nodate} on binary quantization of DNNs. The overall approach consists of using the quantized version of the network during forward and backward computations done throughout training, while performing updates on a full precision copy of the parameters (\emph{e.g.} weights and biases). A critical ingredient of these methods is the use of a so-called Straight-Through Estimator (STE)~\cite{bengio_estimating_2013} that allows the backpropagation of gradients with respect to quantized binary variables (weights and activation signals). Subsequent work~\cite{zhou_dorefa-net_2018} extends these ideas to larger bit widths and tackles quantization of gradient signals as well. Other methods propose learning parameters that bound the range of activation signals inside the network~\cite{choipact2018} and new gradient estimates that allow learning of appropriate scaling factors for integer-based quantization of both weights and activations~\cite{esser2019learned}.

\section{Methodology}
\subsection{Floating-point Encoding}
The floating-point formats we use, denoted as E$e$M$m$, are based on the IEEE-754 standard~\cite{8766229}. A floating-point value $X$ with $m$ bits of mantissa and $e$ bits of exponent is represented (in binary notation) as

\[
  X = {(-1)}^{s}\times 1.\underbrace{x_1\ldots x_m}_{M_X}\times 2^{E_X-E_B},
\]
where $s$ is the sign bit, $M_X$ is the $m$-bit fractional mantissa ($M_X\in[0,1)$) and $E_X\in[0, 2^e-1]$ is the integer exponent. $E_B$ is the exponent bias term, which in the case of IEEE-like encodings is $E_B=2^{e-1}-1$. With respect to a fully IEEE-compliant format, we do not support $\pm\infty$, \texttt{NaN} encodings, and subnormal values. This leads to more representable values and simpler hardware (see~\cite{WP530}). Values that overflow are saturated to the maximal representable number, whereas zeros are represented by $E_X=0$ and $M_X=0$. The values that would have been treated as subnormals ($E_X=0, M_X>0$ and $X=(-1)^s\times 0.M_x\times 2^{-E_B}$) are instead viewed as one extra binade of normal values.

The actual exponent $E_X-E_B\in[-2^{e-1}+1,2^{e-1}]$, as described in the IEEE-754 standard, covers an almost symmetric range of positive and negative values. If the dynamic range of the data to quantize is known, $E_B$ can be adapted to better match it. The PTQ \texttt{AdaptivFloat}~\cite{tambe2020algorithm} method does this by examining the maximum magnitude of the weight tensors in each layer of the network. In a QAT setting, $E_B$ can be better chosen by learning it using a STE~\cite{bengio_estimating_2013} approach. Just as for the scaling factor of an integer-based format, a similar method~\cite{kuzmin2022fp8} can be used to learn a real bias exponent in the floating-point case. In order to achieve a more hardware-friendly minifloat format, we propose to further quantize this learned real exponent bias to an integer. We explore this in Sec.~\ref{sec:fp_qat}.

\subsection{Quantization Scheme}\label{sec:fp_qat}
Our work can be seen as an extension of~\cite{kuzmin2022fp8}. While initially considered for FP8 quantization with subnormals and a real exponent bias, we investigate its use in a sub $8$-bit FP context without subnormals and an integer exponent bias. Subnormal support adds some hardware overhead, but for minifloat formats with small mantissa ($2$ or $3$ bits), using the subnormal range as normal values can still lead to good results with a much smaller overhead (see~\cite{tatsumi2022mixing}).


We quantize weights and activations using an STE-based approach as follows:
\begin{align*}
    \textbf{Forward: } & \mathbf{X}_q = \texttt{quantize}\left(\mathbf{X}, E_0\right) \\
    \textbf{Backward: } &  \dfrac{\partial \ell}{\partial \mathbf{X}} = \dfrac{\partial \ell}{\partial \mathbf{X}_q}\dfrac{\partial \mathbf{X}_q}{\partial \mathbf{X}}
\end{align*}
where $\mathbf{X}$ is an unquantized weight or activation signal, $\ell$ is the loss function, and $\mathbf{X}_q$ is the E$e$M$m$ quantized version of $\mathbf{X}$ computed using Algorithm~\ref{algo:1}. The straight-through moniker comes from the fact that we take $\partial\mathbf{X}_q/\partial\mathbf{X}=1$.

We propose to learn the exponent biases of each layer's weights and activations during training. The learned real values are rounded up (line 1 of Algorithm~\ref{algo:1}) to scale the network signals (weights \& activations) by powers of two. The real values are stored for future use in the update phase with a SGD-type procedure (going from an iteration $t$ to $t+1$) as summarized in Algorithm~\ref{algo:2} (lines 18 and 19). The $g$ quantities in lines 8--13 represent the gradients of the loss $\ell$ with respect to the activation, weight and exponents, respectively. Their computation (the \texttt{backward} function calls) is handled through the PyTorch autograd engine. A schematic view of the entire QAT iteration is also given in Figure~\ref{fig:quant}.

\begin{algorithm}[t]
\small
\caption{\texttt{quantize}: minifloat quantization algorithm}
\label{algo:1}
\begin{algorithmic}[1]\onehalfspacing
\REQUIRE real-valued tensor $\mathbf{X}$, floating-point format E$e$M$m$, real learned exponent bias $E_{0}$.
\ENSURE quantized tensor $\mathbf{X}_{q}$ in the E$e$M$m$ format.

\STATE $E_B = \left\lceil E_{0}\right\rceil$\\
\STATE $x_{\min} = 2^{-E_B} \cdot (1 + 2^{-m})$\\
\STATE $x_{\max} = 2^{2^{e}-1-E_B} \cdot (2 - 2^{-m}$)\\
\STATE $\mathbf{X}_\text{c}=\texttt{clamp}\left(\mathbf{X}, -x_{\max}, x_{\max}\right)$
\STATE $\mathbf{X}_\text{scale} = 2^{\left\lfloor \log_2\left(\mathbf{X_\text{c}}\right)\right\rfloor -m}$
\STATE $\mathbf{X}_\text{q} =\left\lfloor\dfrac{\mathbf{X}_\text{c}}{\mathbf{X}_\text{scale}}\right\rceil\cdot \mathbf{X}_\text{scale}$
\vspace{0.4mm}
\STATE $\mathbf{X}_\text{q} = \mathbf{X}_\text{q}\cdot \mathbb{I}_{|\mathbf{X}_\text{q}|\geq x_{\min}}$
\RETURN $\mathbf{X}_q$
\end{algorithmic}
\end{algorithm}

\begin{algorithm}
\small
\caption{FP QAT algorithm for training a $L$-layer model}
\label{algo:2}
\begin{algorithmic}[1]\onehalfspacing
\REQUIRE a minibatch of inputs $\mathbf{Y}^t_0$ and targets $\mathbf{T}^t$, weights $\mathbf{W}^t\in\mathbb{R}$, activation exponent biases $E^t_{0,a}$, weight exponent biases $E^t_{0,w}$, learning rate $\eta>0$, loss function $\ell$.
\ENSURE updated parameters $\mathbf{W}^{t+1}$, $E^{t+1}_{0,w}$ and $E^{t+1}_{0,a}$

\textbf{\STATE 1. Forward propagation:}
\FOR{$k = 1 \; \textbf{to} \; L$}
\STATE $\mathbf{W}^t_{q,k} \leftarrow \texttt{quantize}\left(\mathbf{W}^t_k, E^t_{0,w,k}\right)$
\STATE $\mathbf{\Tilde{Y}}^t_k \leftarrow \texttt{forward}\left(\mathbf{Y}^t_{q, k-1}, \mathbf{W}^t_{q,k}\right)$
\STATE $\mathbf{Y}^t_{k} \leftarrow \texttt{quantize}\left(\mathbf{\Tilde{Y}}^t_k, E^t_{0,a} \right)$
\ENDFOR
\STATE \textbf{2. Backward propagation:}
\STATE $g_{\mathbf{Y}^t_L} = \dfrac{\partial \ell\left(\mathbf{Y}^t_L, \mathbf{T}^t \right)}{\partial \mathbf{Y}^t_L}$
\FOR{$k = L\; \textbf{down to} \; 1$}
\STATE $g_{\mathbf{Y}^t_{k-1}} \leftarrow \texttt{backward\_activ}\left(g_{\mathbf{Y}^t_{q,k}}, \mathbf{W}^t_{q,k} \right)$
\STATE $g_{\mathbf{W}^t_k} \leftarrow \texttt{backward\_weight}\left(g_{\mathbf{Y}^t_{q,k}}, \mathbf{Y}^t_{q,k} \right)$
\STATE $g_{E^t_{0,a,k}} \leftarrow \texttt{backward\_exp\_bias\_a}\left(g_{\mathbf{Y}^t_{q,k}}, \mathbf{Y}^t_{q,k} \right)$
\STATE $g_{E^t_{0,w,k}} \leftarrow \texttt{backward\_exp\_bias\_w}\left(g_{\mathbf{Y}^t_{q,k}}, \mathbf{Y}^t_{q,k} \right)$
\ENDFOR

\STATE \textbf{3. Parameter update:}
\FOR{$k = 1 \; \textbf{to} \; L$}
\STATE $\mathbf{W}^{t+1}_{k} \leftarrow \texttt{update}\left(\mathbf{W}^t_k, g_{\mathbf{W}^t_k}, \eta\right)$
\STATE $E^{t+1}_{0,w,k} \leftarrow \texttt{update}\left(E^t_{0,w,k}, g_{E^t_{0,w,k}}, \eta\right)$
\STATE $E^{t+1}_{0,a,k} \leftarrow \texttt{update}\left(E^t_{0,a,k}, g_{E^t_{0,a,k}}, \eta\right)$
\ENDFOR
\end{algorithmic}
\end{algorithm}

\section{Experiments}
We apply our approach on a image classification problem on the CIFAR-10 dataset~\cite{krizhevsky2010cifar} and on a satellite image segmentation task using a lightweight U-Net model (adapted from~\cite{vaze2020low}).

\begin{figure*}[h]
    \centering
    \includegraphics[width=0.75\textwidth]{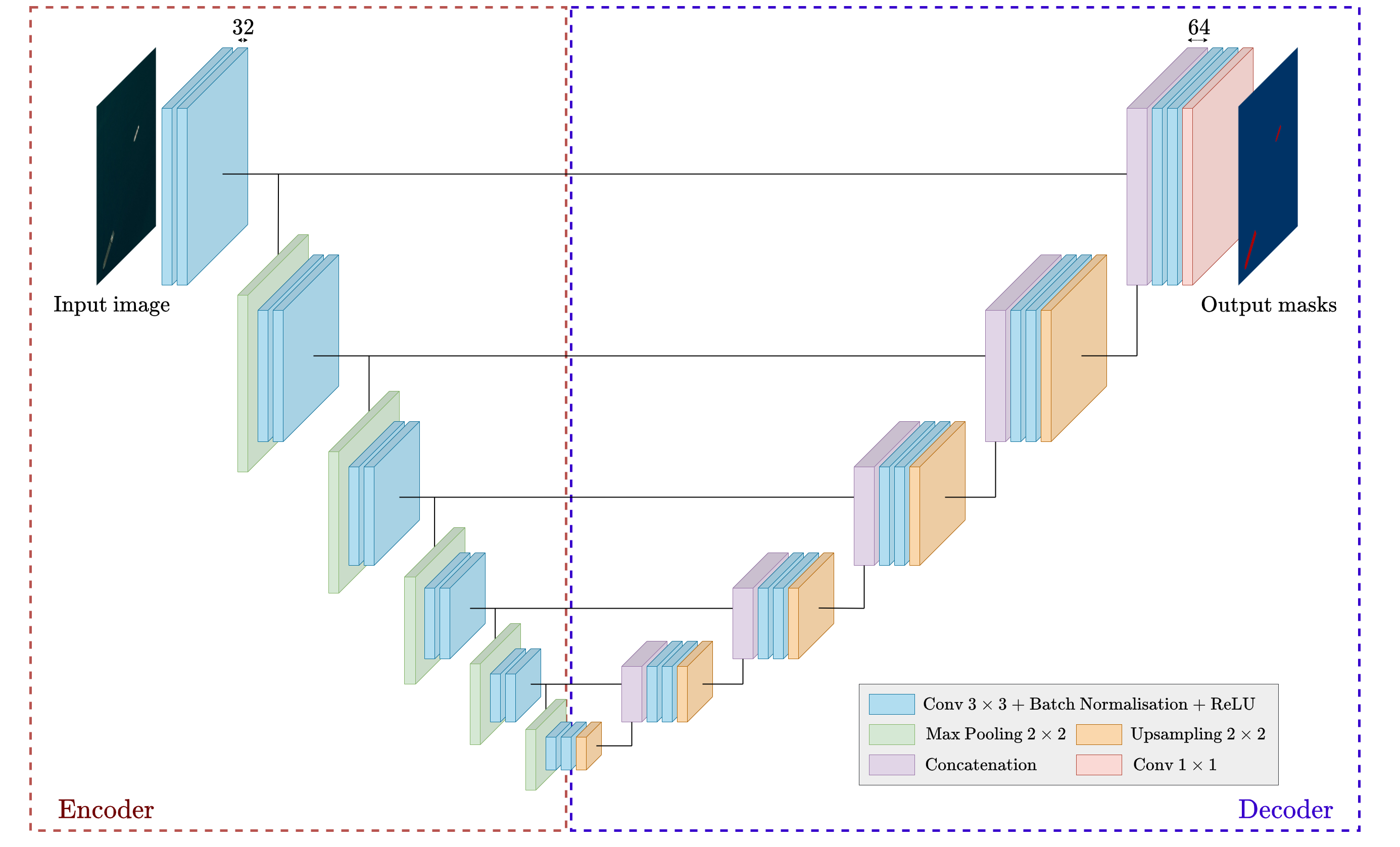}
    \caption{Thin U-Net 32 architecture. It consists of a $5$-stage encoder, followed by a $5$-stage decoder, with skip connections between each corresponding stage pairs in the encoder-decoder blocks.}
    \label{fig:model}
\end{figure*}

\subsection{CIFAR-10 Dataset and Implementation Details}
The CIFAR-10 dataset consists of $32\times32$ pixels RGB labeled images divided into $10$ categories. The dataset is composed of $50,000$ training images and $10,000$ test images. We trained a ResNet-20 model~\cite{he2016deep} from scratch for $300$ epochs using a cross entropy loss function. We apply the data augmentation operations proposed in~\cite{lee2015deeply} over the training set and we use an SGD optimizer with $128$ batch size, weight decay set to $0.0001$, and momentum to $0.9$. Weights are initialized using the Kaiming method~\cite{he2015delving} and the learning rate is scheduled using cosine annealing with starting rate set to $0.1$.

\subsection{Airbus Ship Dataset and Implementation Details}
The Airbus Ship Dataset\footnote{Kaggle Airbus Ship Detection Challenge (July 2018): \url{https://www.kaggle.com/c/airbus-ship-detection}} consists of $768\times768$ pixels RGB satellite images of ships with sea and harbor in the background. The training data is composed of $192,555$ labelled images, including $42,555$ images with ships and $150,000$ background images. To balance the data, we removed $130,000$ background images and split the remaining $62,555$ images into $80\%$ for training and $20\%$ for testing.

We use a Thin U-Net~\cite{vaze2020low} model (see Fig.~\ref{fig:model}), a smaller version of the larger and more widely known U-Net~\cite{ronneberger2015u} architecture. More specifically, we consider the Thin U-Net $32$ model, where $32$ refers to the number of channels for each convolution. Its memory size is smaller by a factor of $290$ compared to a standard U-Net architecture (going down from $288.08$Mb to $0.99$Mb), with little impact on accuracy. Similar to the standard U-Net, this model consists of two distinct blocks, an encoder and a decoder, interconnected by skip connections. The encoder is composed of $5$ blocks containing two groups of $3\times3$ convolution + batch normalisation + ReLU with $2\times2$ maxpooling to down sample the feature maps. Followed by the decoder, it also has $5$ blocks containing a $2\times2$ bilinear up-sampling layer and two groups of $3\times3$ convolution + batch normalisation + ReLU. Prediction masks are generated by a final $1\times1$ convolution without padding.

We apply random horizontal and vertical flip, random crop to resize images to $256\times256$ pixels, random brightness and random contrast as data augmentations on the training set. We use ADAM as the optimizer with a batch sizes of $32$ and an aggregated global loss consisting of a Jaccard loss combined with binary cross entropy with logits, which according to~\cite{wang2020hard} gives the best results in practice.
The learning rate is initially set to $0.001$ and we use a multi step scheduler in order to reduce the learning rate by $0.5$ every $200$ epochs. We train the network for $600$ epochs, using the Kaiming weight initialization method~\cite{he2015delving}. For computing the quantized model, we used the QAT approach from Algorithm~\ref{algo:2} to fine tune the pretrained single-precision model for $50$ epochs. 

Just like with an integer scaling factor~\cite{bhalgat2020lsq+}, a good initialization of the exponent bias parameter is key to convergence with good accuracy. We have found that in practice good initial estimates can be determined by first training the network for a small number of iterations without optimizing the exponent bias (we did this for $200$ iterations in our tests) and then picking it based on the values with maximum magnitude in the weight and activation tensors seen during this process, leading to
\[
    E_0 = 2^{e-1} - \left\lceil\log_2\left(\frac{\max|\mathbf{X}|}{2-2^{-m}}\right)\right\rceil.
\]

\begin{table}[t]
\caption{Comparison of prediction accuracy for CIFAR-10 with different arithmetic formats and bit-widths for weights (W) and activations (A).}
\begin{center}
    \begin{tabular}{ c  c  c  c }
    \hline
    Arithmetic Format & Top-1 & W bit-width & A bit-width\\ \hline
    Single precision & 92.5 & M23E8 & M23E8 \\ \hline
    Fixed-point & 90.2 & 3 & 3 \\
    Integer & 91.6 & 3 & 3 \\ \hline
    \multirow{3}{*}{Minifloat} & 91.3 & M1E1 & M2E2 \\
    & 91.3 & M1E2 & M1E2 \\
    & 90.8 & M1E1 & M1E2 \\
    \hline
    \end{tabular}
\end{center}
\label{tab:2}
\end{table}

\begin{table*}[t]
\caption{Comparison of prediction accuracy for the Thin U-Net 32 model on the Airbus Ship dataset with different arithmetic formats and bit-widths.}
\begin{center}
    \begin{tabular}{ c  c  c  c  c  c}
    \hline
    Format & mean IoU & W bit-width & A bit-width & scaling factor & zero encoding\\ \hline
    Single-precision & 71.0 & M23E8 & M23E8 & / & $M_X=0$ and $E_X=0$ \\\hline
    Fixed-point & 44.5 & 6 & 6 & $2^{\left\lceil\log_2\left(\max|X|\right)\right\rceil}$  & Zero point $= 0$\\
    \multirow{2}{*}{Integer} & 70.5 & 6 & 5 & learn & Zero point $= 0$ \\
     & 68.3 & 5 & 4 & learn & Zero point $= 0$ \\\hline
    \textbf{\multirow{8}{*}{Minifloat}} & 63.4 & E3M2 & E3M2 & $2^{2^{e-1}}$ & $E_X=0$\\
     & 64.8 & E3M2 & E3M2 & $2^{2^{e-1}}$ &  $M_X=0$ and $E_X=0$\\ \cline{2-6}
     & 70.1 & E3M3 & E3M3 & learn & $E_X=0$\\
     & 70.0 & E3M2 & E3M2 & learn &  $E_X=0$\\ \cline{2-6}
     & 71.4 & E4M2 & E4M2 & learn & $M_X=0$ and $E_X=0$\\
     & 70.9 & E3M3 & E3M3 & learn &  $M_X=0$ and $E_X=0$\\
     & \textbf{70.7} & \textbf{E3M2} & \textbf{E3M2} & \textbf{learn} & $\mathbf{M_X=0}$ \textbf{and} $\mathbf{E_X=0}$\\
     & 68.1 & E2M2 & E2M2 & learn &  $M_X=0$ and $E_X=0$\\
    \hline
    \end{tabular}
\end{center}
\label{tab:3}
\end{table*}

\subsection{Results}
We use PyTorch 1.13 on a cluster of eight NVIDIA V100 GPUs to perform our experiments. Unlike the majority of integer-based quantization methods described in the literature which quantize the first and last layers to $8$ bits (which is usually a larger word length than that for the other layers), we use the same small length format for all layers. This can lead to slightly better compression ratios, without affecting accuracy.

We have compared our custom floating-point formats with fixed-point and integer arithmetic alternatives, two quantization formats commonly used to accelerate inference. Results on the CIFAR-10 dataset are given in Table~\ref{tab:2}. While fixed-point quantization seems to degrade model accuracy significantly, low-precision floating-point variants are competitive with integer-based alternatives for very low quantization levels.

The results of applying minifloat quantization to ship detection are summarized in Table~\ref{tab:3}. On this more complex task, 6-bit floating-point formats are necessary to match single precision accuracy, which is evaluated using an \emph{Intersection over Union} (IoU) metric.

\begin{figure*}[h]
\begin{multicols}{4}
\centering
\begin{center}
\includegraphics[width=0.2\textwidth]{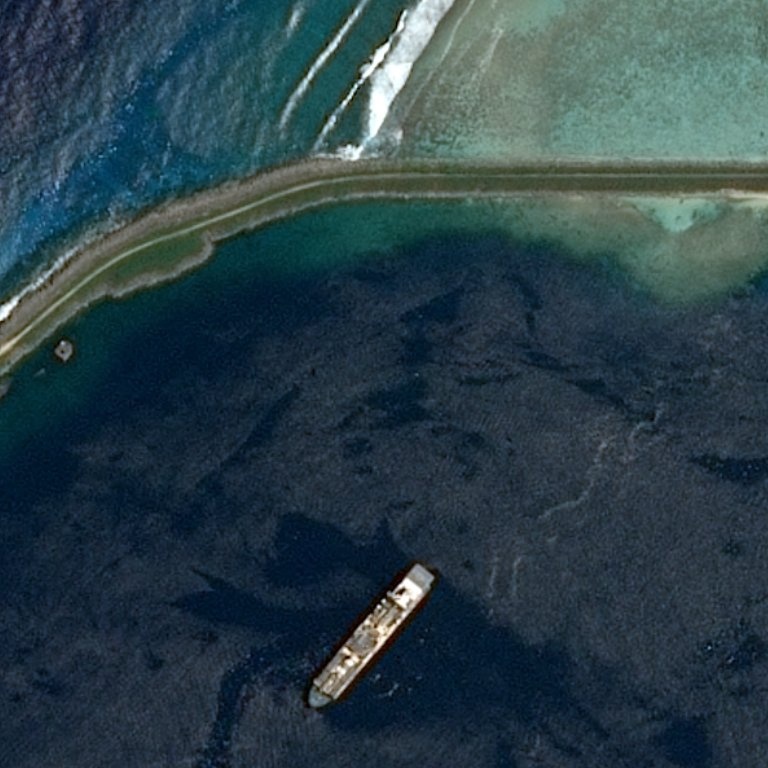}\\
Original image
\end{center}
\begin{center}
\includegraphics[width=0.2\textwidth]{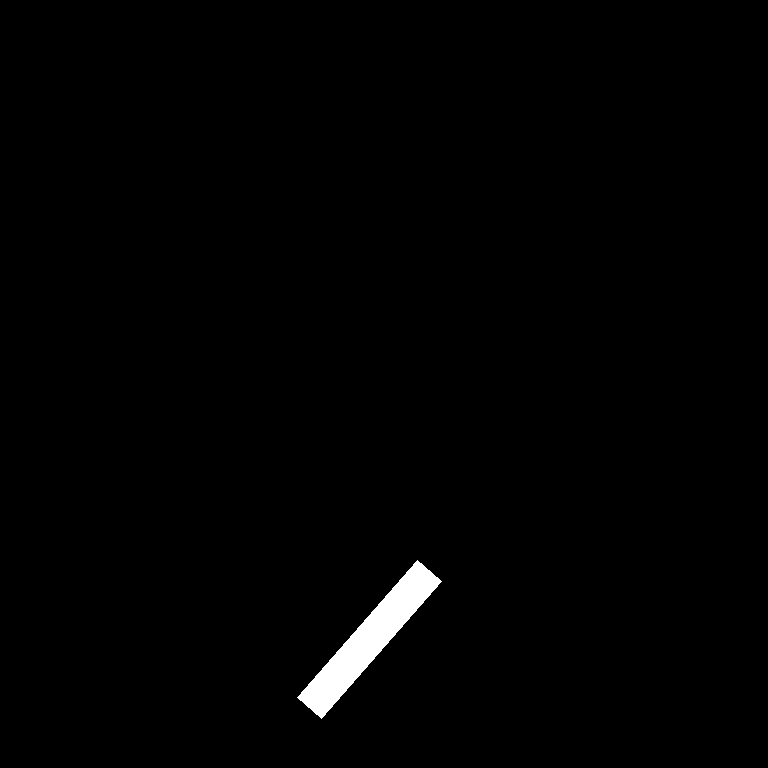}\\
Ground truth
\end{center}
\begin{center}
\includegraphics[width=0.2\textwidth]{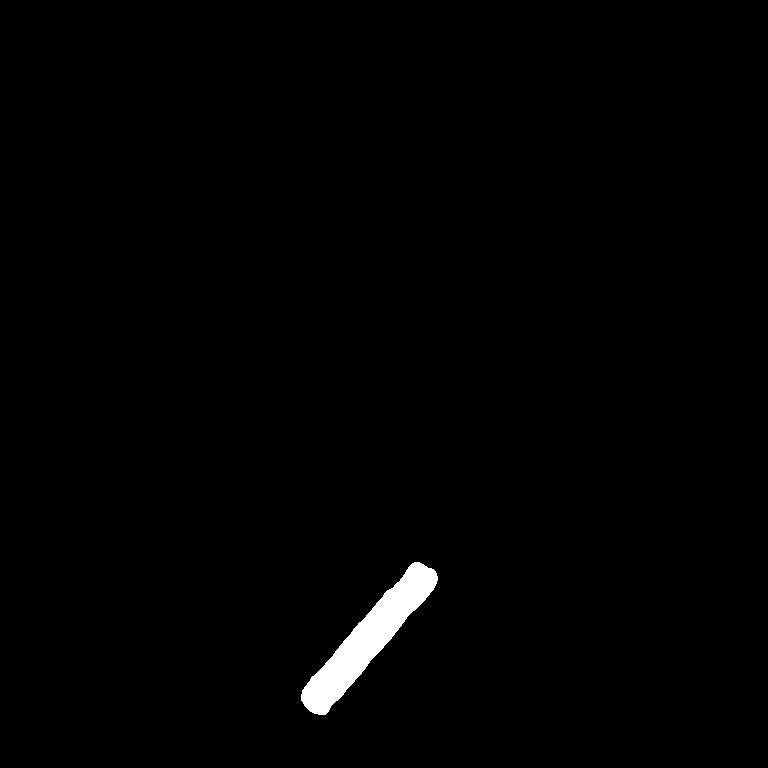}\\
Single-precision
\end{center}
\begin{center}
\includegraphics[width=0.2\textwidth]{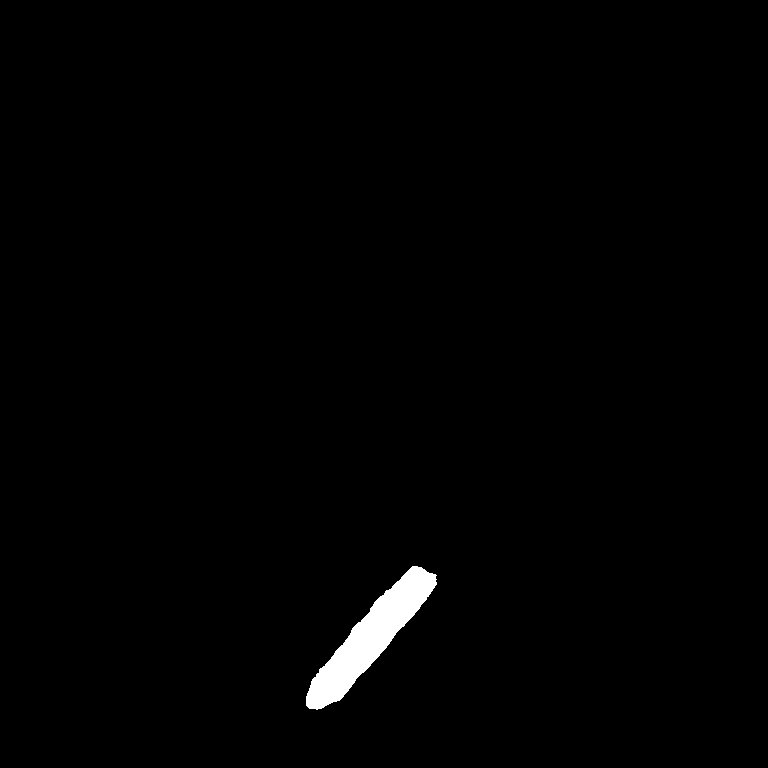}\\
E3M2 minifloat
\end{center}
\end{multicols}
\caption{Prediction of a single ship with original image and its associated masks.}
\label{img:single}

\begin{multicols}{4}
\centering
\begin{center}
\includegraphics[width=0.2\textwidth]{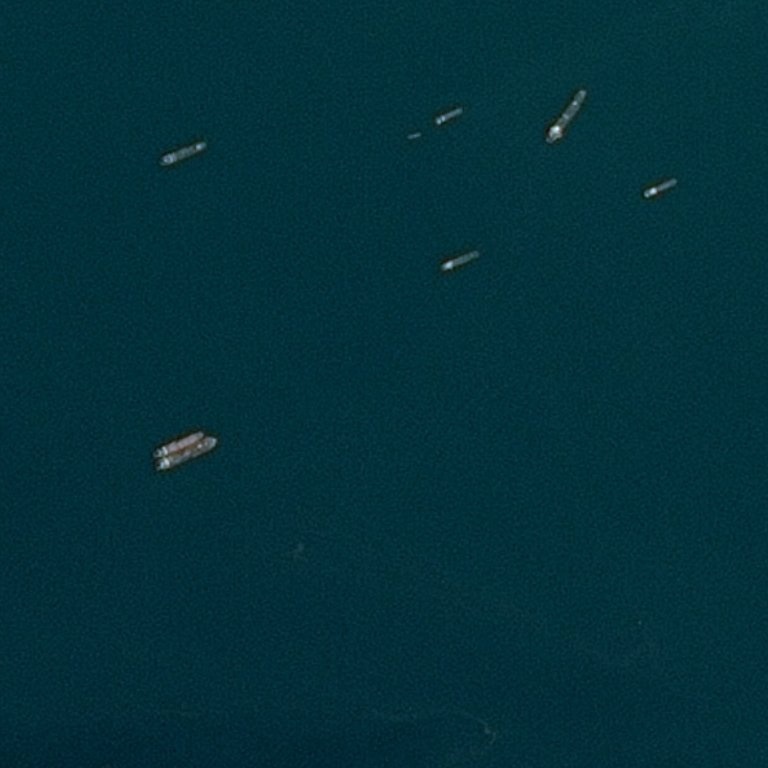}\\
Original image
\end{center}
\begin{center}
\includegraphics[width=0.2\textwidth]{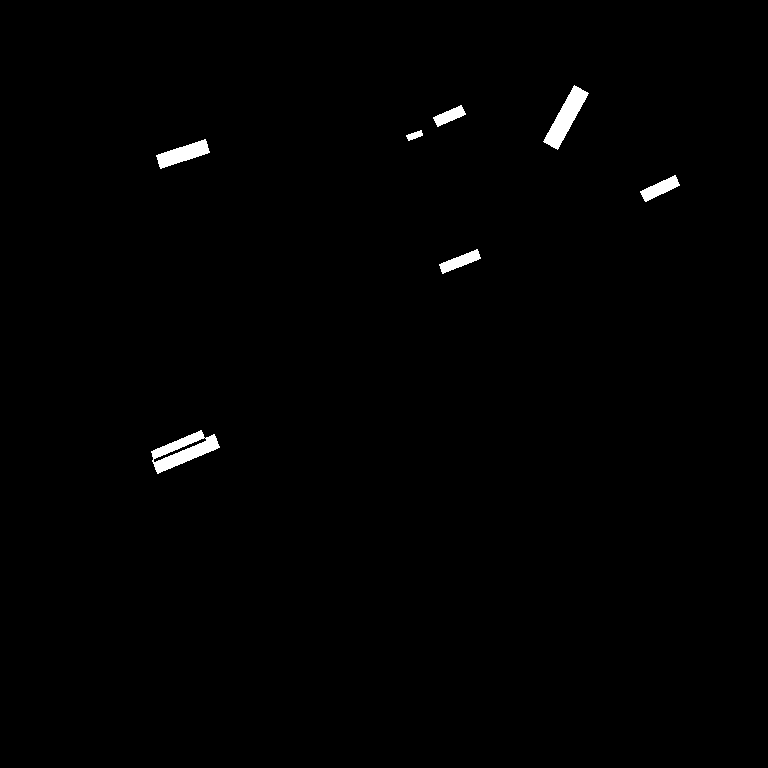}\\
Ground truth
\end{center}
\begin{center}
\includegraphics[width=0.2\textwidth]{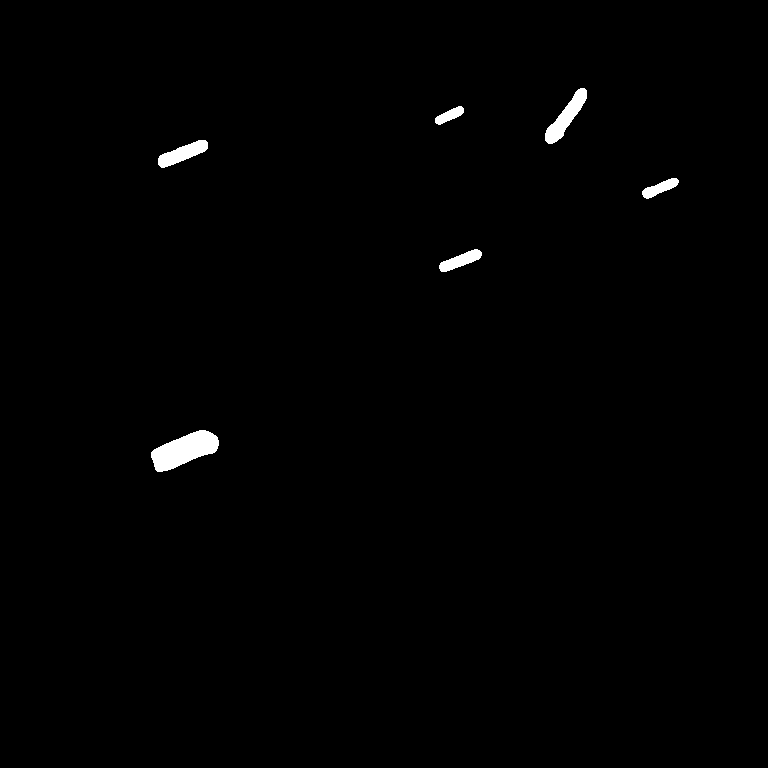}\\
Single-precision
\end{center}
\begin{center}
\includegraphics[width=0.2\textwidth]{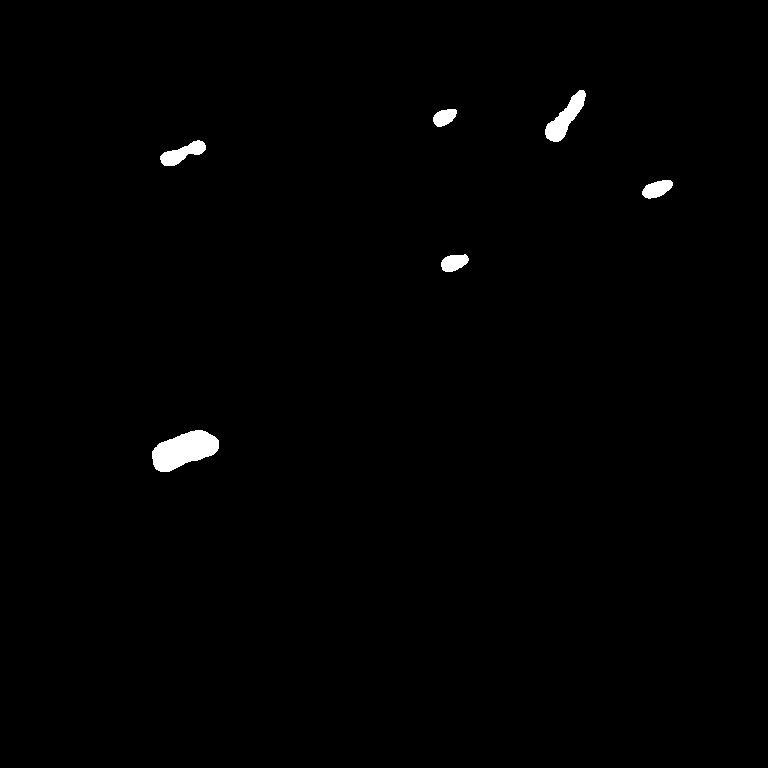}\\
E3M2 minifloat
\end{center}
\end{multicols}
\caption{Prediction of multiple ships and side-by-side ships with original image and its associated masks.}
\label{img:multi}

\begin{multicols}{4}
\centering
\begin{center}
\includegraphics[width=0.2\textwidth]{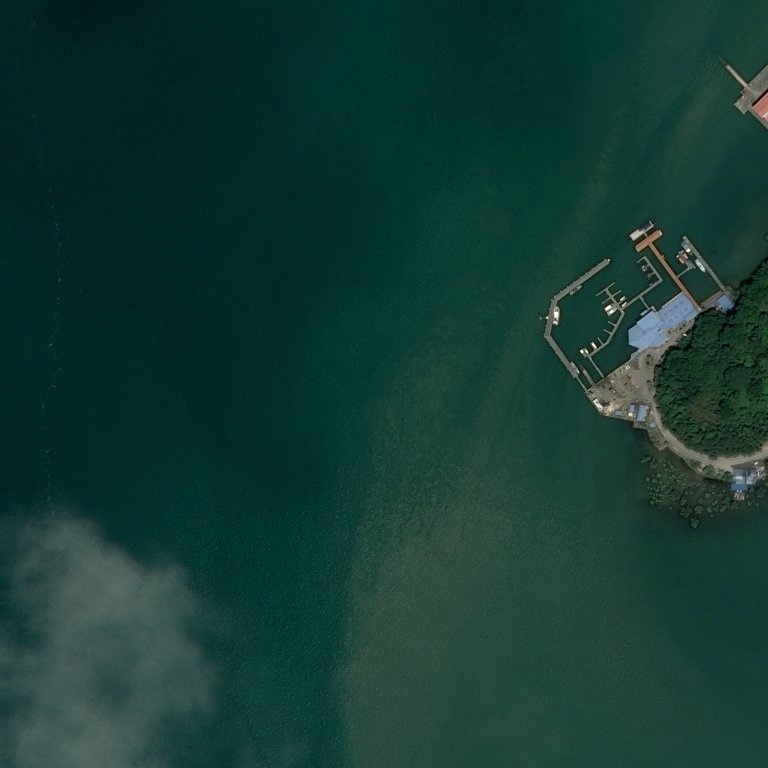}\\
Original image
\end{center}
\begin{center}
\includegraphics[width=0.2\textwidth]{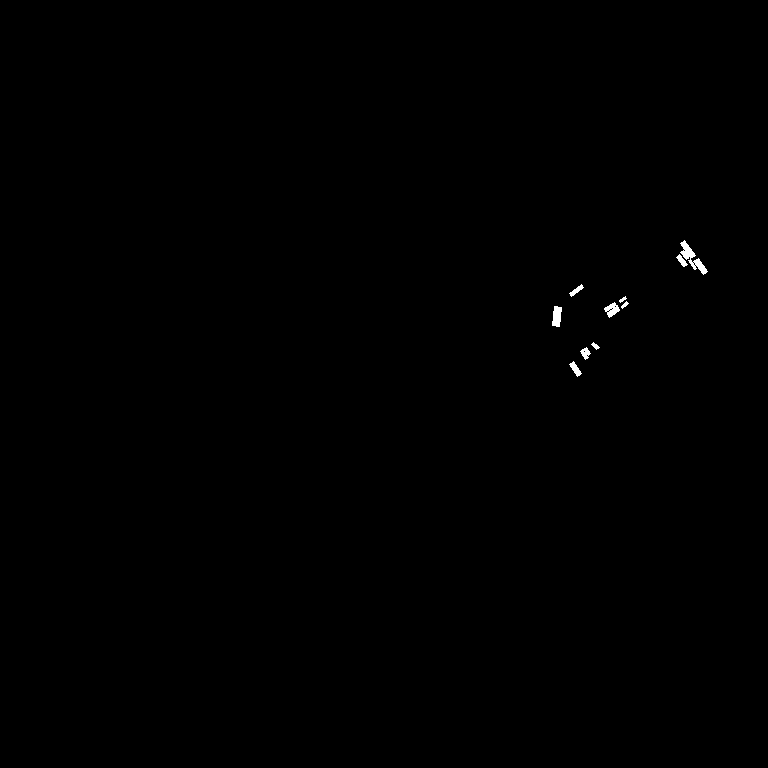}\\
Ground truth
\end{center}
\begin{center}
\includegraphics[width=0.2\textwidth]{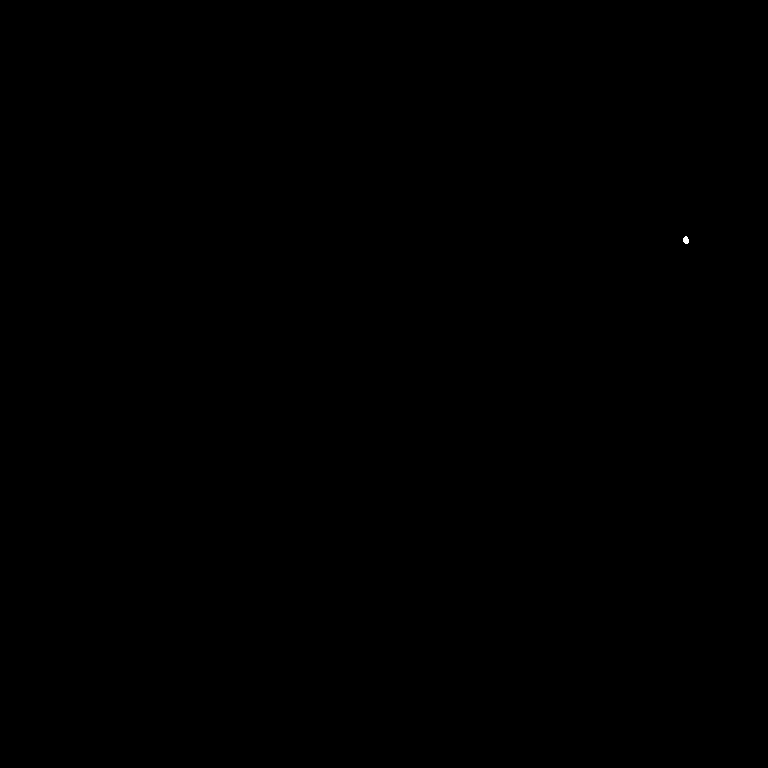}\\
Single-precision
\end{center}
\begin{center}
\includegraphics[width=0.2\textwidth]{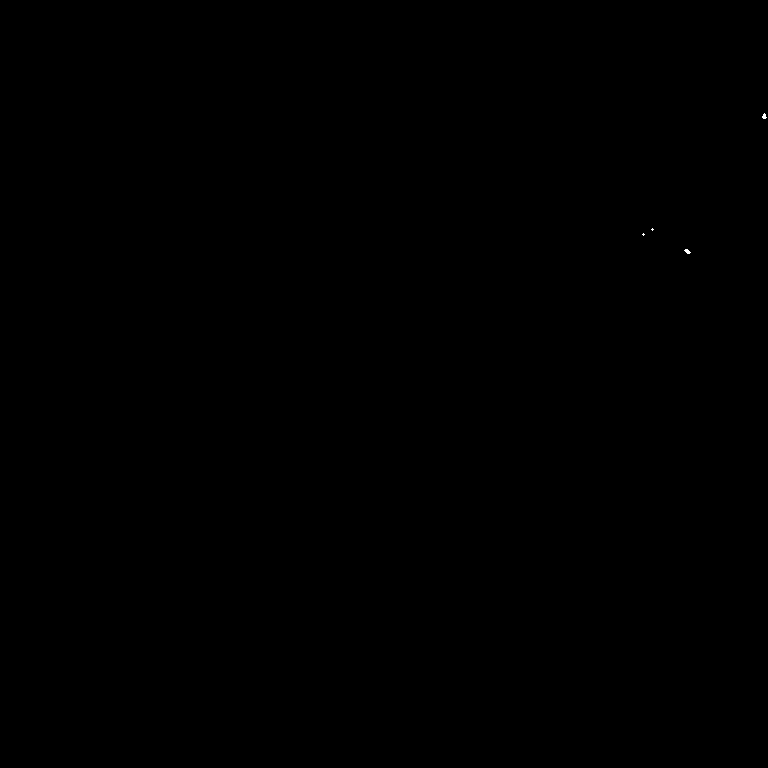}\\
E3M2 minifloat
\end{center}
\end{multicols}
\caption{Small ships in harbor with original image and its associated masks.}
\label{img:harbor}
\end{figure*}

We note that a visual analysis of prediction outputs for the minifloat quantized model shows results that are relatively close to the single-precision baseline. When single ships are present in the image, our quantized model usually does a good job of detecting them (Fig.~\ref{img:single}). Detecting small ships, as well as side by side ships and inshore ships is more challenging than detecting single large ones, leading to poorer predictions even with single-precision models (Fig.~\ref{img:multi} and Fig.~\ref{img:harbor}). Inshore~\cite{nie2018inshore} and small ship~\cite{zhang2019r} detection are challenging topics, both subject to active research.

\subsection{Hardware Implementation Aspects}

The results we have shown so far suggest that minifloat quantization is potentially a good choice for low-precision inference acceleration. However, floating-point addition is in general more resource-intensive than its integer/fixed-point counterpart. 


This is somewhat counterbalanced by the multiplier, which in case of minifloats can be implemented efficiently using just look-up tables (LUTs), as opposed to an 8-bit integer variant that would require DSP blocks, which are much lower in number than LUTs on modern FPGAs (\emph{e.g.} in a AMD-Xilinx UltraScale+ VUP13 FPGA, for every
DSP block there are 140 6-input LUTs~\cite{WP530}).
DSPs can then be configured to implement adder trees for the accumulation part of multiply-accumulate (MAC) units, improving overall logic density.
For instance, the results presented by AMD-Xilinx in~\cite{WP530} claim 60\% higher performance and 12.5\% memory traffic reduction (leading also to lower power usage) when using a minifloat E$3$M$3$ format as opposed to INT8 in a ResNet-50 accelerator. To achieve this performance, they implemented as basic building block a hybrid MAC operator that combines the best of both worlds, a LUT-based minifloat multiplier and a simpler fixed-point adder.

\begin{figure}[h]
    \centering
    \includegraphics[width=0.45\textwidth]{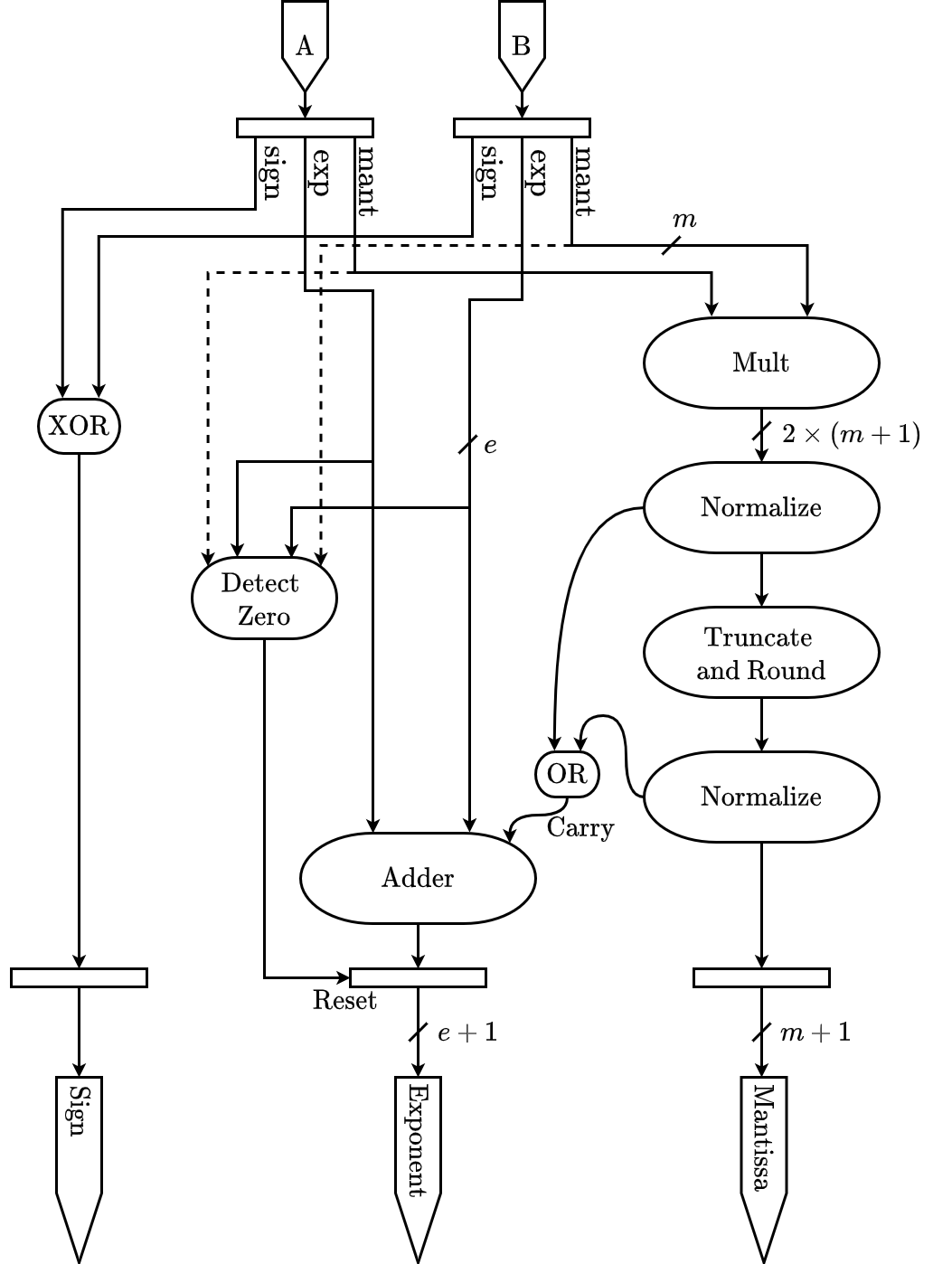}
    \caption{Implementation of a minifloat E$e$M$m$ multiplier (adapted from~\cite{WP530}). The minifloat multiplier preserves exponent accuracy by using a $(e+1)$-bit exponent output and truncates the mantissa output to $m+1$ bits, reducing the size of the logic needed to convert minifloat to fixed point in a hybrid minifloat fixed-point MAC design. To implement the $E_X=0, M_X=0$ zero encoding at the hardware level, we add the input mantissas as inputs to the block \textit{Detect Zero} (dashed lines). This slightly complicates the zero detection logic compared to the $E_X=0$ design from~\cite{WP530}.}
    \label{fig:mul}
\end{figure}

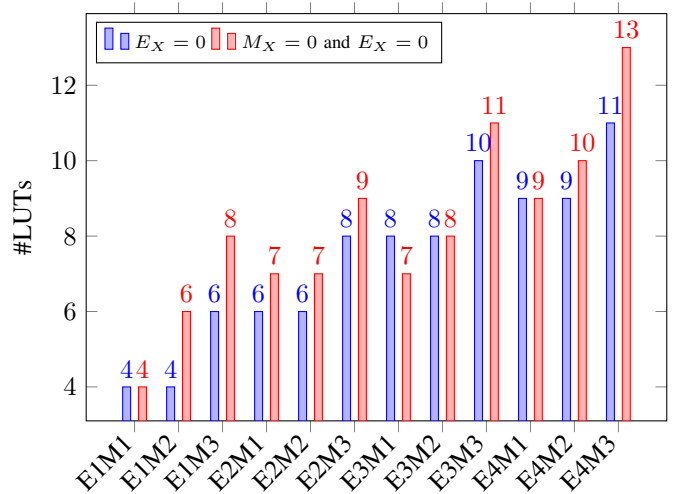
\begin{figure}[H]
\begin{tikzpicture}
\begin{axis}[
    ybar,
    width=9.3cm,
    height=7cm,
    bar width=3,
    enlargelimits=0.1,
    legend style={at={(0.315,0.98)}, font=\fontsize{7}{5}\selectfont, anchor=north,legend columns=-1},
    y label style={at={(axis description cs:0.05,.5)}},
    ylabel={\#LUTs},
    symbolic x coords={E1M1, E1M2, E1M3, E2M1, E2M2, E2M3, E3M1, E3M2, E3M3, E4M1, E4M2, E4M3},
    xtick=data,
    xticklabel style={rotate=45,anchor=east},
    nodes near coords,
    nodes near coords align={vertical},
    ]
\addplot+[bar shift=-3pt] coordinates {(E1M1,4) (E1M2,4) (E1M3,6) (E2M1,6) (E2M2,6) (E2M3,8) (E3M1,8) (E3M2,8) (E3M3,10) (E4M1,9) (E4M2,9) (E4M3,11)};
\addplot+[bar shift=3pt] coordinates {(E1M1,4) (E1M2,6) (E1M3,8) (E2M1,7) (E2M2,7) (E2M3,9) (E3M1,7) (E3M2,8) (E3M3,11) (E4M1,9) (E4M2,10) (E4M3,13)};
\legend{$E_X=0$, $M_X=0$ and $E_X=0$}
\end{axis}
\end{tikzpicture}
    \caption{LUT consumption when implementing a minifloat multiplier with two different zero encodings.}
    \label{fig:lut_count}
\end{figure}


Compared to our minifloat formats where zeros are encoded with $M_X=0$ and $E_X=0$, the formats from~\cite{WP530} are slightly different, opting for a larger range of zero code words corresponding to $E_X=0$. While the AMD-Xilinx choice leads to a slightly more efficient minifloat multiplier design (see Fig.~\ref{fig:mul} for a schematic view of the multiplier and an explanation), our synthesis results using Verilog and Vivado 2022.1 with a Zynq UltraScale+ ZCU102 as target show that the differences in LUT count between the two zero encoding choices are modest (see Fig.~\ref{fig:lut_count}). The extra encoding space saved by our choice ($M_X\neq 0$) has a positive impact on accuracy, as can be seen Table~\ref{tab:3} for the E3M2 and E3M3 formats. We believe that these properties make the $M_X=0$ and $E_X=0$ encoding a better choice in practice. 


The AMD-Xilinx ResNet-50 implementation uses a real-valued scaling factor. With an integer exponent bias like we propose, the logic needed to handle its propagation (\emph{e.g.}~in convolution operations) would amount to a simpler integer exponent shift. In the case of a hybrid MAC design, this could also potentially lead to simpler logic when converting from integers (accumulator outputs) to minifloat (multiplier operands).
We leave the testing of these statements as future work.



\section{Conclusion \& On-going work}
Data transmission from satellites remains challenging due to the many limitations of downlink communication. While DNNs are successfully applied to spatial data processing, deep learning algorithms are now being considered as an on-board alternative to extract the relevant data to be sent to ground stations. However, strict hardware limitations of on-board systems make the use of vanilla DNN models in 32-bit floating-point arithmetic unrealistic.

We propose a QAT algorithm for learning compressed low-precision floating-point DNN models. In addition, we learn the exponent biases of each layer for both weights and activations. Our experiments on the CIFAR-10 and Airbus Ship datasets show good results, with low-precision floating-point models being competitive with single precision baselines. 

To show the potential impact of using floating-point data formats, we have also suggested an implementation of a minifloat-enabled multiplier based on~\cite{WP530} that can be used as a basis for a full DNN inference accelerator for our models. The next step will be to design, test and deploy such an accelerator for the full quantized Thin U-Net 32 on FPGA targets to better gauge the feasibility of using deep learning models with low-precision FP data in an on-board space context.

\section*{Acknowledgment}
We would like to thank Stéphane May for his help and advice. This work was performed using HPC resources from GENCI-IDRIS (Grant 2023-AD011013080R1).

\bibliographystyle{abbrv}

\end{document}